\documentclass[sigconf]{acmart}

\AtBeginDocument{%
  \providecommand\BibTeX{{%
    \normalfont B\kern-0.5em{\scshape i\kern-0.25em b}\kern-0.8em\TeX}}}

\usepackage[utf8]{inputenc}

\usepackage{soul}
\usepackage{float}
\usepackage{multirow}
\usepackage{delarray}

\usepackage{amssymb}
\usepackage{graphicx}

\usepackage{balance}

\newcommand{\ie}{\textit{i}.\textit{e}.}
\newcommand{\eg}{\textit{e}.\textit{g}.}

\newcolumntype{C}[1]{>{\centering\arraybackslash}m{#1}}
\newcolumntype{L}[1]{>{\raggedright\arraybackslash}m{#1}}
\newcolumntype{R}[1]{>{\raggedleft\arraybackslash}m{#1}}
\newcolumntype{N}{@{}m{0pt}@{}}

\copyrightyear{2023}
\acmYear{2023}
\setcopyright{acmlicensed}\acmConference[KDD '23]{Proceedings of the 29th ACM SIGKDD Conference on Knowledge Discovery and Data Mining}{August 6--10, 2023}{Long Beach, CA, USA}
\acmBooktitle{Proceedings of the 29th ACM SIGKDD Conference on Knowledge Discovery and Data Mining (KDD '23), August 6--10, 2023, Long Beach, CA, USA}
\acmPrice{15.00}
\acmDOI{10.1145/3580305.3599402}
\acmISBN{979-8-4007-0103-0/23/08}

\begin{document}

\title[LEA: Improving Sentence Similarity Robustness to Typos Using Lexical Attention Bias]{LEA: Improving Sentence Similarity Robustness to Typos Using Lexical Attention Bias}

\author{Mario Almagro}
\authornote{These authors contributed equally to this research.}
\email{mario.almagro@nielseniq.com}
\affiliation{%
  \institution{NielsenIQ Innovation}
  \city{Madrid}
  \country{Spain}
}
\author{Emilio Almazán}
\authornotemark[1]
\email{emilio.almazan@nielseniq.com}
\affiliation{%
  \institution{NielsenIQ Innovation}
  \city{Madrid}
  \country{Spain}
}

\author{Diego Ortego}
\email{diego.ortego@nielseniq.com}
\affiliation{%
  \institution{NielsenIQ Innovation}
  \city{Madrid}
  \country{Spain}
}

\author{David Jiménez}
\authornotemark[1]
\email{david.jimenez@nielseniq.com}
\affiliation{%
  \institution{NielsenIQ Innovation}
  \city{Madrid}
  \country{Spain}
}

\begin{abstract}
Textual noise, such as typos or abbreviations, is a well-known issue that penalizes vanilla Transformers for most downstream tasks. 
We show that this is also the case for sentence similarity, a fundamental task in multiple domains, \eg~matching, retrieval or paraphrasing. 
Sentence similarity can be approached using cross-encoders, where the two sentences are concatenated in the input allowing the model to exploit the inter-relations between them.  
Previous works addressing the noise issue mainly rely on data augmentation strategies, showing improved robustness when dealing with corrupted samples that are similar to the ones used for training. 
However, all these methods still suffer from the token distribution shift induced by typos.
In this work, we propose to tackle textual noise by equipping cross-encoders with a novel LExical-aware Attention module (LEA) that incorporates lexical similarities between words in both sentences. By using raw text similarities, our approach avoids the tokenization shift problem obtaining improved robustness.
We demonstrate that the attention bias introduced by LEA helps cross-encoders to tackle complex scenarios with textual noise, specially in domains with short-text descriptions and limited context. 
Experiments using three popular Transformer encoders in five e-commerce datasets for product matching show that LEA consistently boosts performance under the presence of noise, while remaining competitive on the original (clean) splits.
We also evaluate our approach in two datasets for textual entailment and paraphrasing showing that LEA is robust to typos in domains with longer sentences and more natural context.
Additionally, we thoroughly analyze several design choices in our approach, providing insights about the impact of the decisions made and fostering future research in cross-encoders dealing with typos.
\end{abstract}




\received{02 February 2023}

\maketitle

\section{Introduction}
\label{sec1:intro}
The fast pace of information systems in society makes noise to be present in almost every text generated by either humans or automated processes. 
Medical reports, queries to databases, messages in social media, receipt transcriptions or product titles in e-commerce are a few examples where real production systems need to cope with a high presence of textual noise like typos, misspellings or custom abbreviations.

\begin{figure}[htp] 
    \includegraphics[width=0.91\linewidth]{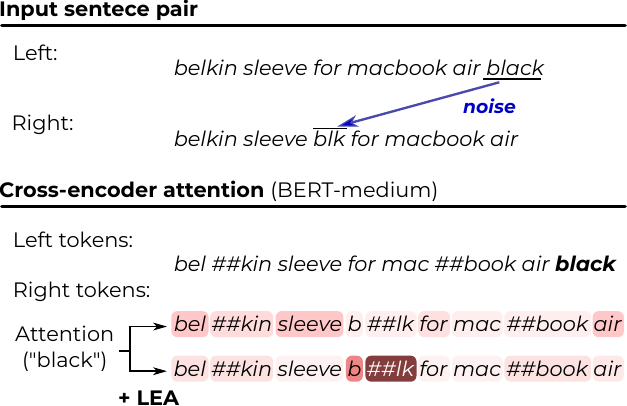}
    \caption{Influence of LEA's lexical bias in the overall attention of one sample in Abt-Buy (BERT-medium cross-encoder). We show the attention of one word on the left sentence (\textit{black}) to all tokens on the right one. Note that the stronger the red color is, the higher the attention.}
    \label{fig:graphical_example}
\end{figure}

Sentence similarity is of great interest to the scientific community for its variety of downstream applications (\eg~question-answering, matching or retrieval) and the unresolved challenges that arises \cite{2016_arXiv_MSMARCO, 2022_SIGIR_BlockSCL, 2021_KDD_BaiduRetrievalSearch}.
Transformer architectures dominate the state-of-the-art with two main alternatives: bi-encoders \cite{2019_EMNLP_SBERT, 2021_EMNLP_SimCSE, 2022_NAACL_diffcse, 2022_SIGIR_BlockSCL, 2022_arXiv_SGPT} and cross-encoders \cite{2019_arXiv_monoBERT, 2020_arXiv_PARADE_CrossReRank, 2020_ACLW_BERT_PM, 2022_arXiv_zeroShotCrossEncRetrieval, 2022_EMNLP_VIRT}. 
Bi-encoders focus on learning meaningful representations for each sentence independently using Siamese-like architectures, making them suitable for efficient retrieval~\cite{2021_ICLR_TransEncoder}.
However, these type of models rely only on the sentence embeddings comparison and lack of any interaction between the tokens/words of the two sentences to be compared.
Cross-encoders \cite{2022_arXiv_zeroShotCrossEncRetrieval}, on the other hand, tackle this limitation by concatenating the two sentences in the input. The attention heads in the Transformer learn the intra and inter-sentence interactions, which in many cases provides highly valuable information for achieving correct predictions in similarity tasks.

Cross-encoders are often considered an upper-bound for textual similarity~\cite{2022_EMNLP_VIRT}, being its main limitation the computational cost of jointly encoding pairs of sentences. 
Recent works, attempt to use its potential by using late interaction modules for bi-encoders~\cite{2020_SIGIR_Colbert,2021_NeurIPS_Baleen,2022_NAACL_Colbertv2},
distillation techniques~\cite{2021_ICLR_TransEncoder, 2022_EMNLP_VIRT}, or using cross-encoders as re-rankers after retrieving potential candidates with bi-encoders~\cite{2019_arXiv_monoBERT, 2020_arXiv_PARADE_CrossReRank}.
These two stages resemble the typical pipeline of product matching, with an initial blocking stage to discard samples that are unlikely to be positive matches for a given product.
This is followed by a more computationally intensive step that identifies the correct match among the selected candidates~\cite{2020_ACLW_BERT_PM, 2020_VLDB_Ditto, 2022_SIGIR_BlockSCL}.

All these methods suffer when dealing with textual noise, that may appear in many forms: \eg~typos and misspellings as in queries for retrieval tasks \cite{2022_SIGIR_AnalysisTypos} or custom abbreviations in certain domains~\cite{2018_BCB_AbbreviationsClinicalDomain, 2020_CLEF_AbbreviationsTweetsSTS}. 
This noise is challenging for vanilla Transformers for two main reasons: character information is not used in this type of architectures, and the shift in the token distribution caused by noise makes harder to relate tokens referring to the same concept, \eg~\textit{chocolate} with \textit{chcolate} or \textit{chclt}. Prior works in information retrieval evidence performance issues~\cite{2022_SIGIR_AnalysisTypos} under these conditions and propose to address them mainly by training with similar typos as the ones in test. Although, this strategy has been proven, to some extent, effective to mitigate the issue of the token distribution shift, all these methods still have the limitations associated to the loss of character information from the tokenization process. 




All the evidence from previous works stress the importance of the character-level information to deal with textual noise.
Following the same intuition, we propose to equip cross-encoders with a LExical Attention bias (LEA) that modifies the self-attention module in Transformers, guiding it towards lexically similar words. This helps the model to improve robustness under the presence of noise. 
We adopt standard data augmentation strategies to deal with typos and demonstrate large performance improvements when adding LEA to cross-encoders.
Figure~\ref{fig:graphical_example} shows an example of the average attention across layers in a cross-encoder. When the typo appears in the word \textit{black} (\ie~\textit{black} $\rightarrow$ \textit{blk}), the tokenizer breaks it into two sub-words (\textit{b} + \textit{\#\#lk}), preventing the self-attention mechanism from capturing the relationship between both terms. 
However, when LEA is introduced, the lexical similarity between both adds a bias that helps the modified attention to maintain this relationship.
Our main contributions are as follows: 

\begin{itemize}
  \item We propose to add a lexical bias in the self-attention module of cross-encoders, which is designed specifically to improve textual noise scenarios.
  This bias provides the model with information about the lexical similarities between words in pairs of sentences, increasing the attention between those that are lexically close.
  
  \item We evaluate LEA in five e-commerce datasets using three Transformer backbones and demonstrate that LEA consistently improves performance by a large margin. Concretely, we report an average improvement of around 6 absolute points of F1-score when dealing with synthetically generated typos. Results in textual entailment and paraphrasing tasks show that LEA achieves competitive performance or even surpasses the baselines.

  \item We thoroughly analyze the impact of the different components of LEA to shed light on the design choices made.

\end{itemize}

\section{Related work}
\label{sec2:previous_works}

\subsection{Sentence similarity}

Determining the degree of similarity between two sentences is a fundamental problem for matching, entailment or paraphrasing tasks and is normally tackled using two type of approaches: bi-encoders and cross-encoders.
Bi-encoders are designed to process each sentence independently, obtaining an embedding for each of them. These models are typically trained with metric learning objectives that pull together the representations of positive pairs while pushing apart those of negative pairs. The authors in~\cite{2021_EMNLP_SimCSE} propose SimCSE, which exploits dropout to generate embeddings that build positive pairs in the unsupervised setup. They also propose a supervised setting where they use textual entailment labels to construct the positive pair. Tracz et al.~\cite{2020_ACLW_BERT_PM} adopt a triplet loss in a supervised setup for product matching. The approach described in~\cite{2022_NAACL_diffcse}, extends SimCSE and propose to learn via equivariant contrastive learning, where representations have to be insensitive to dropout and sensitive to MLM-based word replacement perturbations. Supervised contrastive learning~\cite{2021_NeurIPS_SupContLearn} is also adopted in~\cite{2021_EMNLP_SCL_Text} for sentence similarity in a general domain, while~\cite{2022_arXiv_SupConLossPM, 2022_SIGIR_BlockSCL} apply it for product matching in e-commerce. Another popular method is SBERT~\cite{2019_EMNLP_SBERT}, which addresses both the problem of predicting several similarity degrees as a regression task and directly matching pairs of sentences via classification.

Cross-encoders, on the other hand, jointly process a concatenated pair of sentences.
These models are often considered to outperform bi-encoders~\cite{2021_ICLR_TransEncoder, 2022_EMNLP_VIRT}, obtaining robust results in general domains~\cite{2022_EMNLP_VIRT}, product matching~\cite{2020_VLDB_Ditto} or retrieval tasks~\cite{2022_arXiv_zeroShotCrossEncRetrieval}. 
However, their main drawback is the need to recompute the encoding for each different pair of sentences.
Therefore, many recent works adopt hybrid solutions to improve bi-encoders.
Humeau et al. proposed Poly-encoders~\cite{2020_ICLR_PolyEncoder} that utilizes an attention mechanism to perform extra interaction after Siamese encoders.
The TransEncoder method~\cite{2021_ICLR_TransEncoder} alternates bi- and cross-encoder independent trainings, while distilling their knowledge via pseudo-labels. The resulting bi-encoder shows improved performance.
Distillation is further explored in~\cite{2022_EMNLP_VIRT}, where knowledge transfer from the cross-attention of a light interaction module is adopted during training and removed at inference time.

\subsection{Dealing with typos and abbreviations}
Recent literature demonstrates that Transformer-based architectures \cite{2017_NeiurIPS_AttentionAllNeed} are not robust to textual noise, \ie~to misspellings or abbreviations of the input words ~\cite{2022_SIGIR_CharacterTypos, 2022_SIGIR_AnalysisTypos, 2020_arXiv_AdvBERT, 2021_EMNLP_EvaluatingPerturbations, 2019_WNUT_NoiseTranslation}. 
Despite using sub-word tokenizers (\eg~WordPiece) designed to deal with out-of-vocabulary words, Transformers exhibit in practice performance drops when exposed to typos.
Words with noise are not likely to be present in the vocabulary and therefore, they are split into several sub-words yielding to token distribution shifts with respect to the noise-free counterpart~\cite{2022_SIGIR_CharacterTypos,2021_EMNLP_DealingTypos}.

Training the model with synthetically generated perturbations~\cite{2021_EMNLP_DealingTypos, 2019_WNUT_NoiseTranslation, 2022_SIGIR_AnalysisTypos} is a standard practice to deal with typos.
Some techniques use simple addition, deletion or character swaps, while others use more sophisticated methods such as common misspellings or keyboard and OCR errors~\cite{2021_EMNLP_DealingTypos}. 
Moreover, depending on the type of perturbation, the same type of noise can have a different impact, \ie~typos in various words of a sentence do not influence equally. This issue was reported in~\cite{2022_SIGIR_AnalysisTypos, 2020_arXiv_AdvBERT} showing that noise in relevant words yields larger performance drops.
We can find approaches that complement this practice with different architectural designs or specific losses. 
For example, in~\cite{2022_SIGIR_CharacterTypos} the authors realize that character-based Transformers provide improved robustness to typos and exploit this fact to propose a self-teaching strategy to boost performance of dense retrievers. Authors in~\cite{2019_ACL_CombatingMisWordRecogn} add a module to recognize the presence of typos in words just before the downstream classifier, which helps to align representations between noisy and clean versions.

\subsection{Relative attention bias}
Self-attention modules in Transformers receive as input the token representations coming from the previous layers and output contextual representations for each token estimated from a weighted combination of the tokens' representations. 
Modifications of the self-attention have been proposed to add a bias that accounts for the relative distance between words/tokens in the input sequence ~\cite{2018_NAACL_RelativePosition, 2019_CORR_transXL, 2020_JMLR_relPosExpl, 2021_NEURIPS_relposcape}. This strategy known as relative positional embeddings replaces the absolute positional embeddings, where the position was injected as part of the input of the Transformer.
In~\cite{2021_NAACL_RelPosLongShort} the authors follow this idea and extend it with long and short term relations. 
Wennberg et al.~\cite{2021_ACL_RelPosTransFree} propose a more interpretable representation for translation invariant relative positional embeddings using the Toeplitz matrix.
The authors in~\cite{2022_ICLR_ALIBI} simplify the embeddings by just adding fixed scalars to the attention values that vary with the distance.
This way of adding relative information between tokens has also been applied to information extraction from documents, where they use 2D relative distances~\cite{2022_AAAI_BROS} or tabular structural biases for table understanding~\cite{2022_ACL_TableFormer}.

\section{Proposed approach}
\label{sec3:method}

In this work we propose to incorporate a Lexical-aware Attention module (LEA) to the self-attention mechanism of vanilla cross-encoder Transformers. 
This module considers inter-sentence lexical relations, which we demonstrate to be key for improving sentence similarity tasks, specially in the presence of typos.

\paragraph{Notation} We use capital letters to denote sets (\eg~``$X$''), bold capital letters for matrices (\eg~``$\mathbf{X}$''), bold lowercase letters for vectors (\eg~``$\mathbf{x}$'') and lowercase for scalars (\eg~``$x$''). For simplicity in the notation, equations only refer to a single layer and head in the Transformer architecture.

\subsection{Self-attention}
A key cornerstone in the success of Transformers is the multi-head self-attention mechanism, which learns token dependencies and encodes contextual information from the input~\cite{2021_AAAI_SelfAttention}. 
In particular, a single head in this attention module receives an input sequence of $n$ token representations coming from the previous layer $X = (\mathbf{x}_1, \dots, \mathbf{x}_n)$, 
where $\mathbf{x}_i \in \mathbb{R}^{d_{h}}$ and computes a new sequence $Z=\left( \mathbf{z}_1,\dots, \mathbf{z}_n \right)$ of the same length and hidden dimension $d_h$. 
The resulting token representations are computed as follows: 

\begin{equation}
\label{eqn:attention_output}
    \mathbf{z_i} = \displaystyle \sum_{j=1}^{n} a_{ij} \left( \mathbf{x}_j \cdot \mathbf{W}^V \right), \quad  \mathbf{z_i}\in \mathbb{R}^{d_{h}}.
\end{equation}
Therefore, each new token representation $\mathbf{z_i}$ is a weighted average of the linearly projected tokens representations $\mathbf{x}_j$, using the value projection matrix $\mathbf{W}^V$. 
The weight associated with each pair of tokens $a_{ij}$ is computed using a softmax function: 

\begin{equation}
    a_{ij} = \displaystyle \frac{\exp{e_{ij}}}{\sum^n_{k=1} \exp{e_{ik}}},
\end{equation}
where the scalar $e_{ij}$ is computed using a compatibility function (dot product) between tokens $i$ and $j$ in the sentence: 

\begin{equation}
    \label{eq:attention}
    e_{ij} = \displaystyle \frac{(\mathbf{x}_i \mathbf{W}^Q) (\mathbf{x}_j \mathbf{W}^K)}{\sqrt{d_{h}}}.
\end{equation}
The query, key and value projection matrices $\{\mathbf{W}^Q, \mathbf{W}^K, \mathbf{W}^V\} \in \mathbb{R}^{d_h \times d_i}$ are learned during training, where $d_i$ refers to the dimension of the intermediate representations.

\subsection{Lexical attention bias for cross-encoders \label{subsec: LEA}}

\begin{figure}[htb]
    \includegraphics[width=0.7\linewidth]{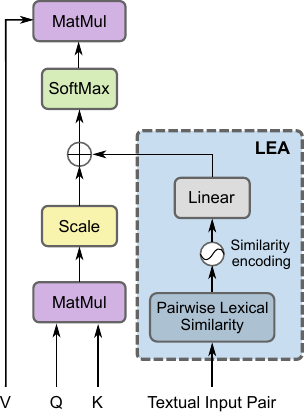}
    \caption{Overview of the attention mechanism in Transformers where we add the proposed lexical attention bias (LEA). We use the traditional nomenclature for the key, query and value representations (Q, K, V).}
    \label{fig:general_diagram}
\end{figure}

As we demonstrate in Section~\ref{sec4:results}, in presence of textual noise, $a_{ij}$ struggles to relate similar terms corrupted by noise.
To address this issue, we propose to add a lexical attention bias to the self-attention module of cross-encoders. 
This bias term guides the attention towards tokens with high lexical similarity. We illustrate our proposed architecture in Figure~\ref{fig:general_diagram}.

Cross-encoders for textual similarity receive as input the concatenation of the two sentence representations to be compared:
\begin{equation}
   \mathbf{X}_c = \mathbf{X}_l ~|~ \mathbf{X}_r,
\end{equation}
where $\mathbf{X}_l$ and $\mathbf{X}_r$ are the left and right sentences, respectively.
Inspired by previous works in relative position embeddings~\cite{2018_NAACL_RelativePosition}, we propose to modify the self-attention module described in Eq.~\ref{eq:attention} as follows:

\begin{equation}
\label{eq:lea_attention}
    \tilde{e}_{ij} = \displaystyle e_{ij} + \alpha~ \boldsymbol{\ell}_{ij} \mathbf{W}^{L}, \quad \forall i,j \in \mathbf{X}_c,
\end{equation}
where the second term accounts for the lexical bias. $\mathbf{W}^L \in \mathbb{R}^{d_{L} \times 1}$ is a learnable projection matrix, 
$\boldsymbol{\ell}_{ij} \in \mathbb{R}^{1 \times d_{L}}$ is the pairwise lexical attention embedding and $\alpha$ is a fixed scale factor that aligns the contributions of the lexical attention ($\boldsymbol{\ell}_{ij} \mathbf{W}^{L}$) and the scaled-dot product attention ($\displaystyle e_{ij}$). This factor is computed automatically once at the beginning of the training based on the magnitudes of both terms. 

To compute the pairwise lexical attention embedding, we first measure the similarity between words considering only inter-sentence relations, \ie~lexical similarities between words of the same sentence are set to $0$:
\begin{equation}
\label{eq:similarity_metric}
s_{ij} = 
\begin{cases}
    \mathit{Sim} \left( w(\mathbf{x}_i), w(\mathbf{x}_j) \right) & \text{, if } \mathbf{x}_i \in \mathbf{X}_l ~and~ \mathbf{x}_j \in \mathbf{X}_r \\
    & ~~~ or ~ \mathbf{x}_i \in \mathbf{X}_r ~and~ \mathbf{x}_j \in \mathbf{X}_l \\
    0 & \text{, otherwise}.
\end{cases}
\end{equation} 
where $\mathbf{X}_l$ and $\mathbf{X}_r$ represent the pair of input sentences to compare, $w(\mathbf{x}_i)$ and $w(\mathbf{x}_j)$ denote the input textual word associated to the i-th and j-th tokens, respectively, and $\mathit{Sim(\cdot, \cdot)}$ is a metric that measures the string similarity between two words. We elaborate on our choice for the similarity metric in Section~\ref{sec4:lexical_metric}.

Inspired by~\cite{2017_NeiurIPS_AttentionAllNeed}, we apply a sinusoidal function over $s_{ij}$ to get an embedding that represents the lexical similarity:

\begin{eqnarray}
\label{eq:lexical_relatt1}
    \boldsymbol{\ell}_{ij}^{(s_{ij},2p)}&=&\sin\left(\frac{2\pi\cdot d_{ij}}{\beta^{2p/d_{h}}}\right),\\
\label{eq:lexical_relatt2}
    & & \nonumber \\
    \boldsymbol{\ell}_{ij}^{(s_{ij},2p+1)}&=&\cos\left(\frac{2\pi\cdot d_{ij}}{\beta^{2p/d_{h}}}\right),
\end{eqnarray}
where $\beta = 10^4$ and $p\in\left\{ 0,\ldots,d_{L}-1\right\}$. The final lexical embedding $\boldsymbol{\ell}_{ij}$ is the concatenation of the two sinusoidal embeddings in Eq.~\ref{eq:lexical_relatt1} and Eq.~\ref{eq:lexical_relatt2}, respectively. 
Different from the original proposal~\cite{2017_NeiurIPS_AttentionAllNeed} we scale the similarity $s_{ij}$ by $2\pi$ to cover the full range of the sinusoidal functions. This results in embeddings more uniformly distributed across the output space.

Note that by equipping LEA with a learnable projection matrix $\mathbf{W}^{L}$ we provide the model with the flexibility to adjust the contribution coming from the lexical term in the final attention values.
The parameter overhead introduced by this term is $d_{L} \times \#\text{heads}$ in all the layers where we use it. 

\section{Experimental work}
\label{sec4:results}
We structure the experimentation around four research questions to shed light on the design and capabilities of LEA.

\vspace{0.15cm}
$\mathbf{RQ_1}$. Does LEA improve performance in a consistent way across datasets and architectures under the presence of typos while remaining competitive on clean setups?

\vspace{0.15cm}
$\mathbf{RQ_2}$. How important is the choice of the lexical similarity metric for LEA under the presence of typos?

\vspace{0.15cm}
$\mathbf{RQ_3}$.What is the impact of applying LEA on varying locations of the architecture and the effect of sharing the parameters at different levels \eg~model, layer?

\vspace{0.15cm}
$\mathbf{RQ_4}$. Does LEA generalize to different noise strengths?

\vspace{0.2cm}
\noindent The remaining of this section presents the experimental setting in Section~\ref{subsec:ExpSetting} and responds the four research questions in Sections~\ref{sec4:robustness},~\ref{sec4:lexical_metric},~\ref{sec4:sharing_strategy} and~\ref{sec4:noise_strength}.

\subsection{Experimental setting\label{subsec:ExpSetting}}

The impact of textual noise in the prediction of the models depends on whether it appears on relevant words or not~\cite{2022_SIGIR_AnalysisTypos}. 
We argue that when the sentences are short the probability of the random noise appearing in relevant words is higher and, therefore, we expect a higher contribution of the lexical attention bias.
Hence, the core of our experiments is conducted in five product matching datasets, where the sentences are short and normally with lack of syntax: Abt-Buy~\cite{2018_SIGMOD_DeepMatcher}, Amazon-Google~\cite{2018_SIGMOD_DeepMatcher} and WDC-Computers (small, medium and large)~\cite{2019_WWWC_WDCdataset}.
Moreover, we validate the contribution of LEA in two related tasks of natural language domain: textual entailment (RTE~\cite{2018_ICLR_GLUE}) and paraphrasing (MRPC~\cite{2018_ICLR_GLUE}).
Details about the datasets are provided in Tables~\ref{table:ecom_datasets} and~\ref{table:sts_datasets}, respectively.
We artificially introduce typos on the aforementioned datasets as described in Section~\ref{sec4:noise_generation}.

{\renewcommand{\arraystretch}{1.1}
\begin{table}[t]
    \centering
    \begin{tabular}{L{0.12\textwidth}L{0.05\textwidth}L{0.05\textwidth}L{0.05\textwidth}C{0.12\textwidth} N}
        \hline
        \multirow{2}{*}{\textbf{Dataset}} & \multicolumn{3}{C{0.19\textwidth}}{\textbf{Size}} & \multirow{2}{*}{\shortstack[l]{\textbf{Average}\\ \textbf{\#words}}} & \tabularnewline
        \cline{2-4}
        & \centering Train & \centering Val & \centering Test & \tabularnewline
        \hline
        \hline
        Abt-Buy     & ~5,743 & 1,916 & 1,916 & ~8.0 & \tabularnewline
        Amaz.-Goog. & ~6,874 & 2,293 & 2,293 & ~6.6  & \tabularnewline
        WDC-Comp.   & & & \tabularnewline
        \qquad - small   & ~2,263 & \quad 567 & 1,100 & 12.1 & 
        \tabularnewline
        \qquad - medium   & ~6,464 & 1,618 & 1,100 & 12.2 & \tabularnewline
        \qquad - large   & 26,640 & 6,659 & 1,100 & 13.4 & \tabularnewline
        \hline
        \tabularnewline
    \end{tabular}
    \caption{Product matching datasets details: train/val/test sizes and the average number of words per sentence.
    \label{table:ecom_datasets}}
\end{table}
}

{\renewcommand{\arraystretch}{1.1}
\begin{table}[t]
    \centering
    \begin{tabular}{L{0.08\textwidth}L{0.05\textwidth}L{0.05\textwidth}L{0.05\textwidth}C{0.08\textwidth} N}
        \hline
        \multirow{2}{*}{\textbf{Dataset}} & \multicolumn{3}{C{0.19\textwidth}}{\textbf{Size}} & \multirow{2}{*}{\shortstack[l]{\textbf{Average}\\ \textbf{\#words}}} & \tabularnewline
        \cline{2-4}
        & \centering Train & \centering Val & \centering Test & \tabularnewline
        \hline
        \hline
        RTE  & 2,241 & 249 & ~ 277 & 43.3 & \tabularnewline
        MRPC & 3,668 & 408 & 1,725 & 22.0 & \tabularnewline
        \hline
        \tabularnewline
    \end{tabular}
    \caption{Textual similarity datasets details: train/val/test sizes and the average number of words per sentence.
    \label{table:sts_datasets}}
\end{table}
}

{\renewcommand{\arraystretch}{1.3}
\begin{table*}[htp]
    \centering
    \begin{tabular}{L{0.12\textwidth} L{0.1\textwidth} L{0.12\textwidth} L{0.1\textwidth} L{0.1\textwidth} L{0.1\textwidth} L{0.01\textwidth} L{0.1\textwidth} N}
        \hline
        \multirow{2}{*}{} & \multirow{2}{*}{\shortstack[l]{\textbf{Abt-Buy} \\ (typo)}} & \multirow{2}{*}{\shortstack[l]{\textbf{Amaz.-Goog.} \\ (typo)}} & \multicolumn{3}{C{0.3\textwidth}}{\textbf{WDC-Computers} (typo)} & \multirow{2}{*}{} & \multirow{2}{*}{\textbf{Average}} & \tabularnewline
        \cline{4-6}
        & & & \centering Small & \centering Medium & \centering Large & & & \tabularnewline
        \hline
        \hline
        Electra-small      & 33.15 $\pm$ 8.0 & 24.61 $\pm$ 5.9 & 35.76 $\pm$ 5.4 & 49.73 $\pm$ 3.8 & 53.60 $\pm$ 1.1 & & 39.37 $\pm$ 4.8 & \tabularnewline
        ~~+ DA             & 69.57 $\pm$ 2.2 & 55.50 $\pm$ 2.5 & 72.92 $\pm$ 2.5 & 76.56 $\pm$ 0.6 & 82.34 $\pm$ 0.9 & & 71.38 $\pm$ 1.7 & \tabularnewline
        ~~+ LEA            & \textbf{74.04 $\pm$ 1.7} & \textbf{68.01 $\pm$ 1.3} & \textbf{75.71 $\pm$ 0.9} & \textbf{79.94 $\pm$ 1.5} & \textbf{86.27 $\pm$ 1.3} & & \textbf{76.79 $\pm$ 1.3} & \tabularnewline
        \hline
        BERT-Med.          & 43.69 $\pm$ 5.4 & 17.33 $\pm$ 2.5 & 54.12 $\pm$ 4.3 & 58.35 $\pm$ 2.6 & 53.25 $\pm$ 3.4 & & 45.35 $\pm$ 3.6 & \tabularnewline
        ~~+ DA             & 67.08 $\pm$ 2.9 & 53.41 $\pm$ 1.7 & 69.17 $\pm$ 1.5 & 76.25 $\pm$ 1.1 & 84.89 $\pm$ 0.9 & & 70.16 $\pm$ 1.6 & \tabularnewline
        ~~+ LEA & \textbf{73.19 $\pm$ 2.0} & \textbf{69.30 $\pm$ 1.2} & \textbf{72.64 $\pm$ 0.9} & \textbf{79.51 $\pm$ 1.3} & \textbf{86.66 $\pm$ 0.8} & & \textbf{76.26 $\pm$ 1.2} & \tabularnewline
        \hline
        BERT-Base       & 57.01 $\pm$ 3.0 & 22.42 $\pm$ 2.7 & 51.29 $\pm$ 6.6 & 58.49 $\pm$ 5.1 & 55.32 $\pm$ 2.6 & & 48.91 $\pm$ 4.0 & \tabularnewline
        ~~+ DA          & 70.60 $\pm$ 3.2 & 53.79 $\pm$ 1.1 & 71.57 $\pm$ 0.8 & 77.55 $\pm$ 2.7 & 84.69 $\pm$ 0.9 & & 71.64 $\pm$ 1.8 & \tabularnewline
        ~~+ LEA         & \textbf{75.97 $\pm$ 1.4} & \textbf{70.41 $\pm$ 0.9} & \textbf{76.85 $\pm$ 1.1} & \textbf{82.10 $\pm$ 1.7} & \textbf{88.07 $\pm$ 1.0} & & \textbf{78.68 $\pm$ 1.2} & \tabularnewline
        \hline
        \tabularnewline
    \end{tabular}
    \caption{Results obtained in five e-commerce datasets for product matching with typos. We report the mean and standard deviation of the F1-score for three seeds during training and three versions of the test splits where we add synthetic noise (nine experiments each).}
    \label{table:results_product_typo}
\end{table*}
}

\subsubsection{Synthetic noise generation}
\label{sec4:noise_generation}
The original datasets mostly contain clean sentences with a low percentage of textual noise.
To evaluate the models' generalization under the presence of noise, we synthetically generate test splits containing a wide range of typos. We apply the strategies followed in previous works~\cite{2022_SIGIR_AnalysisTypos, 2020_arXiv_AdvBERT} using the \textit{nlpaug}\footnote{\url{https://github.com/makcedward/nlpaug}} library for data augmentation. In particular, we consider the following character operations:
\begin{itemize}
    \item Insertion. A random character is inserted in a random position within the word, \eg~\textit{screen} $\rightarrow$ \textit{scree$\mathbf{t}$n}.
    \item Deletion. A random character from the word is removed, \eg~\textit{screen} $\rightarrow$ \textit{sceen}.
    \item Substitution. A random character from the word is replaced by a random character, \eg~\textit{screen} $\rightarrow$ \textit{s$\mathbf{b}$reen}.
    \item Swapping. A random character is swapped with one neighbor character in the word, \eg~\textit{screen} $\rightarrow$ \textit{s$\mathbf{rc}$een}.
    \item Keyboard substitution. A random character from the word is replaced by a close character in the QWERTY keyboard, \eg~\textit{screen} $\rightarrow$ \textit{s$\mathbf{c}$teen}.
\end{itemize}

We modify all sentences in the test splits, where each word has a 20\% chance to be augmented. Only one type of operation is applied to each word, which is chosen randomly among the five options. 
We limit the augmentation to words with more than 3 characters to mitigate the effect of words becoming non-recognizable from the original form \eg~\textit{ace} $\rightarrow$ \textit{ate}.

\subsubsection{Baseline models}
\label{sec4:baselines}

Due to the lack of prior works dealing with textual noise in cross-encoders for sentence similarity tasks, we adopt a benchmarking based on the comparison between three versions of cross-encoders: 1) vanilla, 2) trained with data augmentation (DA) and 3) trained with data augmentation and LEA. 
We adopt 2) as the reference baseline in the literature following the approach of related works in other domains, where they successfully applied data augmentation to deal with typos~\cite{2021_EMNLP_DealingTypos, 2019_WNUT_NoiseTranslation, 2022_SIGIR_AnalysisTypos}.

For data augmentation during training we apply the same configuration as in the synthetic generation of the test splits (see Section~\ref{sec4:noise_generation}) and use a 50\% chance for each sentence to be augmented. 
We use three popular pre-trained language models (PLMs) of varying sizes, \ie~Electra-small~\cite{2020_ICLR_ELECTRA}, BERT-Medium~\cite{2019_arXiv_BERTmedium} and BERT-Base~\cite{2019_NAACL_BERT}.

\subsubsection{Implementation details}
\label{sec:implementation}

In all the experiments we fine-tune the PLMs described in Section~\ref{sec4:baselines} for 30 epochs, using AdamW with a  batch size of 32, an initial learning rate of $5e^{-5}$, a weight decay of $5e^{-5}$ and we apply a cosine annealing scheduler with a warm-up of 1.5 epochs.
For LEA, $\alpha$ is automatically fixed in Eq.~\ref{eq:lea_attention} at the beginning of the training for each layer of the Transformer and leave $\mathbf{W}^L$ being trained independently per head. As similarity metric we use Jaccard (see Section~\ref{sec4:lexical_metric} for more details) and apply the proposed lexical attention bias to half of the last layers in all architectures (see Section~\ref{sec4:sharing_strategy} for a detailed analysis). For more details we refer the reader to Appendix~\ref{sec:appendix_setup_details}.

We use the same training data for all methods and evaluate them on two different test splits, the original (clean) and the corrupted version with typos.
We run three different seeds for the training and create three test splits randomly introducing typos, as their influence may differ depending on the words containing typos.
Thus, we report the resulting mean and standard deviation over three and nine results for the clean and typo experiments, respectively.

The test splits with typos, the binaries of the models and the required material to reproduce results are available in our repository\footnote{\url{https://github.com/m-almagro-cadiz/LEA}}.

{\renewcommand{\arraystretch}{1.3}
\begin{table*}[t]
    \centering
    \begin{tabular}{L{0.12\textwidth} L{0.1\textwidth} L{0.12\textwidth} L{0.1\textwidth} L{0.1\textwidth} L{0.1\textwidth} L{0.01\textwidth} L{0.1\textwidth} N}
        \hline
        \multirow{2}{*}{} & \multirow{2}{*}{\shortstack[l]{\textbf{Abt-Buy}\\ (clean) }} & \multirow{2}{*}{\shortstack[l]{\textbf{Amaz.-Goog.} \\ (clean)}} & \multicolumn{3}{C{0.3\textwidth}}{\textbf{WDC-Computers} (clean)} & \multirow{2}{*}{} & \multirow{2}{*}{\textbf{Average}} & \tabularnewline
        \cline{4-6}
        & & & \centering Small & \centering Medium & \centering Large & & & \tabularnewline
        \hline
        \hline
        Electra-small  & 78.85 $\pm$ 0.8 & \textbf{71.80 $\pm$ 0.9} & \textbf{83.37 $\pm$ 1.8} & \textbf{88.42 $\pm$ 0.4} & \textbf{93.00 $\pm$ 0.9} & & \textbf{83.09 $\pm$ 1.0} & \tabularnewline
        ~~+ DA         & 80.62 $\pm$ 1.1 & 69.83 $\pm$ 3.3 & 79.12 $\pm$ 1.7 & 85.65 $\pm$ 1.0 & 92.19 $\pm$ 0.7 & & 81.48 $\pm$ 1.6 & \tabularnewline
        ~~+ LEA        & \textbf{80.95 $\pm$ 1.6} & 70.94 $\pm$ 2.5 & 78.70 $\pm$ 0.5 & 85.23 $\pm$ 1.2 & 91.34 $\pm$ 0.4 & & 81.43 $\pm$ 1.2 & \tabularnewline
        \hline
        BERT-Med.      & 79.21 $\pm$ 2.7 & 69.06 $\pm$ 1.0 & \textbf{80.78 $\pm$ 1.5} & \textbf{86.91 $\pm$ 1.4} & 90.98 $\pm$ 0.5 & & \textbf{81.62 $\pm$ 1.6} & \tabularnewline
        ~~+ DA         & 77.86 $\pm$ 0.3 & 67.45 $\pm$ 2.8 & 73.35 $\pm$ 0.8 & 81.61 $\pm$ 0.8 & 90.60 $\pm$ 0.9 & & 77.87 $\pm$ 1.2 & \tabularnewline
        ~~+ LEA        & \textbf{82.31 $\pm$ 0.2} & \textbf{73.39 $\pm$ 0.9} & 74.66 $\pm$ 0.9 & 83.76 $\pm$ 1.8 & \textbf{91.23 $\pm$ 0.5} & & 79.82 $\pm$ 1.1 & \tabularnewline
        \hline
        BERT-Base      & \textbf{83.03 $\pm$ 0.8} & 70.82 $\pm$ 0.6 & \textbf{82.08 $\pm$ 0.1} & \textbf{88.13 $\pm$ 1.1} & \textbf{92.69 $\pm$ 0.5} & & \textbf{83.35 $\pm$ 0.6} & \tabularnewline
        ~~+ DA         & 80.43 $\pm$ 1.7 & 67.80 $\pm$ 1.6 & 75.83 $\pm$ 1.9 & 84.59 $\pm$ 1.1 & 89.51 $\pm$ 0.9 & & 79.63 $\pm$ 1.4 & \tabularnewline
        ~~+ LEA        & 82.66 $\pm$ 0.9 & \textbf{72.62 $\pm$ 0.4} & 79.14 $\pm$ 1.8 & 86.40 $\pm$ 0.3 & 91.04 $\pm$ 0.7 & & 82.37 $\pm$ 0.8 & \tabularnewline
        \hline
        \tabularnewline
    \end{tabular}
    \caption{Results obtained in five e-commerce datasets for product matching (original splits). We report the mean and standard deviation of the F1-score for three seeds during training.}
    \label{table:results_product_clean}
\end{table*}
}

{\renewcommand{\arraystretch}{1.2}
\begin{table*}[t]
    \centering
    \begin{tabular}{L{0.1\textwidth} L{0.1\textwidth} L{0.1\textwidth} L{0.01\textwidth} L{0.1\textwidth} L{0.1\textwidth} N}
        \hline
        \multirow{2}{*}{} & \multicolumn{2}{C{0.2\textwidth}}{\textbf{RTE}} & & \multicolumn{2}{C{0.2\textwidth}}{\textbf{MRPC}} & \tabularnewline
        \cline{2-3}
        \cline{5-6}
         & \centering Typo & \centering Clean & & \centering Typo & \centering Clean & \tabularnewline
        \hline
        \hline
        BERT-Med. & 19.13 $\pm$ 3.5 & \textbf{70.19 $\pm$ 2.0} & & 34.52 $\pm$ 15.5 & \textbf{86.58 $\pm$ 0.1} &  \tabularnewline
        ~~+ DA    & 59.31 $\pm$ 1.9 & 65.96 $\pm$ 3.3 & & \textbf{82.87 $\pm$ 0.8} & 86.21 $\pm$ 0.3 & \tabularnewline
        ~~+ LEA   & \textbf{65.18 $\pm$ 2.4} & 67.13 $\pm$ 2.0 & & \textbf{82.16 $\pm$ 1.6} & 84.32 $\pm$ 1.5 & \tabularnewline
        \hline
        \tabularnewline
    \end{tabular}
    \caption{Results obtained in two additional domains of sentence similarity, \ie~textual entailment (RTE) and paraphrasing (MRPC).
    We report the mean and standard deviation of the F1-score values for the noisy (nine seeds) and the original (three seeds) versions of the test splits, respectively.}
    \label{table:results_rte_mrpc}
\end{table*}
}

\subsection{Robustness across datasets}
\label{sec4:robustness}

We compare in Table~\ref{table:results_product_typo} the F1-score of LEA with that of vanilla cross-encoders trained without and with data augmentation (+ DA) in five product matching datasets. 
We observe that applying data augmentation to mimic typos during training improves the robustness to them as reported by previous works in the retrieval domain~\cite{2022_SIGIR_AnalysisTypos}.
When we apply LEA, we outperform the baseline by 5.4, 6.1 and 7.0 points on average across the five datasets for Electra-small, BERT-Medium and BERT-Base, respectively.
Strategies solely based on data augmentation completely depend on the tokenized data, which may lose part of the lexical information when splitting into sub-words.
In contrast, LEA exploits character-level similarity between words, an information that is not dependent on the tokenization.

Moreover, in Table~\ref{table:results_product_clean} we analyze the impact of adding LEA to cross-encoders in the absence of typos.
Here, the vanilla cross-encoders trained without data augmentation perform best on average. 
LEA, however clearly outperforms training with data augmentation and provides a competitive performance, achieving the best performance in some datasets.
We refer the reader to Sections~\ref{sec:appendix_exp_larger_models} and \ref{sec:appendix_exp_larger_datasets} for additional experiments with a larger architecture (BERT-Large), autoregressive models (GPT-2 and GPT-Neo) and larger datasets (WDC-XLarge and WDC-All).

The results presented in Tables~\ref{table:results_product_typo} and~\ref{table:results_product_clean}, therefore provide a positive response to $\mathbf{RQ_1}$: \textit{LEA improves cross-encoders performance to typos by a large margin, while achieving competitive performance in their absence}.


\subsubsection{Performance on additional domains}
Previous experiments showing the improvements of LEA were conducted in the e-commerce domain, \ie~short product descriptions with little context. 
In Table~\ref{table:results_rte_mrpc}, we further demonstrate the benefits of LEA using BERT-Medium in RTE (textual entailment) and MRPC (paraphrasing) datasets that represent a completely different domain with longer sentences. 
Again, typos dramatically reduce the performance of a cross-encoder trained without data augmentation. 
However, LEA palliates this drop and achieves best results in RTE with typos ($\sim$ 6 absolute points gain), while having comparable performance to a vanilla cross-encoder trained with data augmentation in MRPC.
In contrast, in a clean setup LEA suffers small performance drops with respect to the cross-encoder. We argue that Jaccard may reflect similarity worse in long texts than an edit distance because it is agnostic to character order, resulting in a higher probability of highlighting unrelated words.
In turn, longer sentences reduce the probability of applying typos to relevant words, thus hiding the potential benefit of using LEA in real settings.
Despite these limitations, we show that even in this situation LEA performs competitively and can even improve performance.

\subsection{Impact of the lexical similarity choice}
\label{sec4:lexical_metric}

The lexical embeddings of LEA are computed with a similarity function between two strings.  
In Table~\ref{table:ablation} (\textbf{Lexical similarity metric}), we analyze the impact of the choice of this similarity metric in the Abt-Buy dataset using BERT-Medium. 
We try LEA with the following string similarity metrics: Jaccard (Jac.), Smith-Waterman (Smith), Longest Common Subsequence (LCS), Levenshtein (Lev.) and Jaro–Winkler (Jaro)~\cite{2019_ESA_framework}. 
All the metrics improve the performance when evaluating with typos, thus supporting the positive contribution of LEA regardless of the lexical similarity metric adopted. 
In clean scenarios, the Smith-Waterman similarity does not outperform the regular cross-encoder (top row), while the remaining metrics does surpass it. 
Smith-Waterman is the metric that is penalized the most by typos appearing in the middle of words, and by lexical variations, as it relies on aligning common substrings.

{\renewcommand{\arraystretch}{1.2}
\begin{table}[htp]  
    \centering
    \begin{tabular}{L{0.18\textwidth} L{0.05\textwidth} L{0.05\textwidth} N}
        \hline
        \textbf{Model} & \textbf{Typo} & \textbf{Clean} & \tabularnewline
        \hline
        \hline
        BERT-Med. (8 layers) & 43.69 & 79.21 & \tabularnewline
        BERT-Med. + DA       & 67.08 & 77.86 & \tabularnewline
        \hline
        \hline
        \multicolumn{3}{C{0.28\textwidth}}{\textbf{Lexical similarity metric \{ $\cdot$ \} }} & \tabularnewline[1.0ex]
        ~~+ LEA \{Smith\}  & 68.29 & 78.95 & \tabularnewline
        ~~+ LEA \{LCS\}    & 71.82 & 80.48 & \tabularnewline
        ~~+ LEA \{Lev.\}   & 71.52 & 80.36 & \tabularnewline
        ~~+ LEA \{Jaro\}   & 72.96 & 81.23 & \tabularnewline
        ~~\textbf{+ LEA \{Jaccard\}} & \textbf{73.19} & \textbf{82.31} & \tabularnewline
        \hline
        \multicolumn{3}{C{0.28\textwidth}}{$\mathbf{W}^L$ \textbf{parameter sharing} ( $\cdot$ ) } & \tabularnewline[1.0ex]
        ~~+ LEA (model) & 70.79 & 80.13 & \tabularnewline
        ~~+ LEA (layer) & 71.56 & 81.71 & \tabularnewline
        ~~\textbf{+ LEA (head)} & \textbf{73.19} & \textbf{82.31} & \tabularnewline
        \hline
        \multicolumn{3}{C{0.28\textwidth}}{\textbf{Layers with LEA [ $\cdot$ ] }} & \tabularnewline[1.0ex]
        ~~+ LEA [0-8] & 68.49 & 75.87 & \tabularnewline
        ~~+ LEA [2-8] & 69.75 & 78.14 & \tabularnewline
        ~~\textbf{+ LEA [4-8]} & \textbf{73.19} & \textbf{82.31} & \tabularnewline
        ~~+ LEA [6-8] & 72.81 & 80.75 & \tabularnewline
        \hline
        \tabularnewline
    \end{tabular}
    \caption{Influence of the different design choices in LEA, \ie~the lexical similarity metric \{Smith-Waterman, Longest Common Subsequence, Levenshtein, Jaro-Winkler and Jaccard\}, the sharing strategy of LEA's projection matrix (shared per model, layer or head) and the layers where LEA is applied within a 8-layer BERT architecture [1-8].}
    \label{table:ablation}
\end{table}
}

We decided to adopt the Jaccard similarity for LEA given that it consistently outperforms both the clean and the noisy scenarios for short sentences. 
The Jaccard coefficient applied to characters is order agnostic and therefore more robust to character swaps.
Our intuition is that Jaccard provides higher separability between word pairs with and without typos, which is beneficial in short-texts. 
However, as the word context increases in long sentence domains, the probability of comparing words with different meaning that share characters increases, thus reducing the swap invariance advantage.
We refer the reader to Appendix~\ref{sec:appendix_relative_bias} for further details on the design choices for the relative attention bias used in LEA.

The evidences presented provide a positive answer to $\mathbf{RQ_2}$: \textit{it is important to choose the right metric for better performance, although all of them help in preventing performance drops against typos with respect to the vanilla cross-encoder}.

\subsection{LEA on different layers and sharing strategy}
\label{sec4:sharing_strategy}
Two important decisions to make when integrating LEA in a Transfomer architecture are:
\begin{itemize}
    \item Do we use the same lexical projection matrix for the entire model, one per layer, or one independent matrix per head?
    \item Do we apply LEA in all layers across the architecture, or it is more beneficial to apply it only in certain layers?
\end{itemize}
In Table~\ref{table:ablation} we present results to answer these questions. 
For the first decision ($\mathbf{W}^L$ \textbf{parameter sharing}) we show that using an independent projection matrix per head behaves best and observe an increasing performance tendency towards sharing less parameters, \ie~shared across all layers is the worst choice. 
We argue that this behaviour is reasonable given that using independent $\mathbf{W}^{L}$ matrices provides higher flexibility to learn the projection, as the addition to the standard self-attention term in Eq.~\ref{eq:lea_attention} might need different behaviour in different heads for better performance. 
We, therefore, use for the default LEA configuration this non-shared alternative.

Regarding the second decision (\textbf{Layers with LEA}), we evaluate adding LEA to different layer subsets in BERT-Medium (8 layers in total): all layers ([0-8]), excluding the first two layers ([2-8]), second half of the layers ([4-8]) and the last two layers ([6-8]). 
We observe that all the choices help dealing with typos, achieving the best performance by adding LEA to the second half of layers.
Similar behaviour is observed in clean scenarios, although only adding LEA to the last half and last two layers outperform the vanilla cross-encoder performance.
Therefore, we use LEA in half of the deeper layers in all architectures and experiments.

We argue that the character-level similarity provided by LEA can be considered as a high-level interaction information. 
Therefore, this complements the high-level features of deep Transformer layers.
We left for future work to validate this hypothesis.

The results obtained in this set of experiments address $\mathbf{RQ_3}$: \textit{it is better to use dedicated lexical projection matrices for each attention head and to add LEA on late layers for better performance}.

\subsection{Impact of the noise strength}
\label{sec4:noise_strength}
\begin{figure}[t]
    \includegraphics[width=\linewidth]{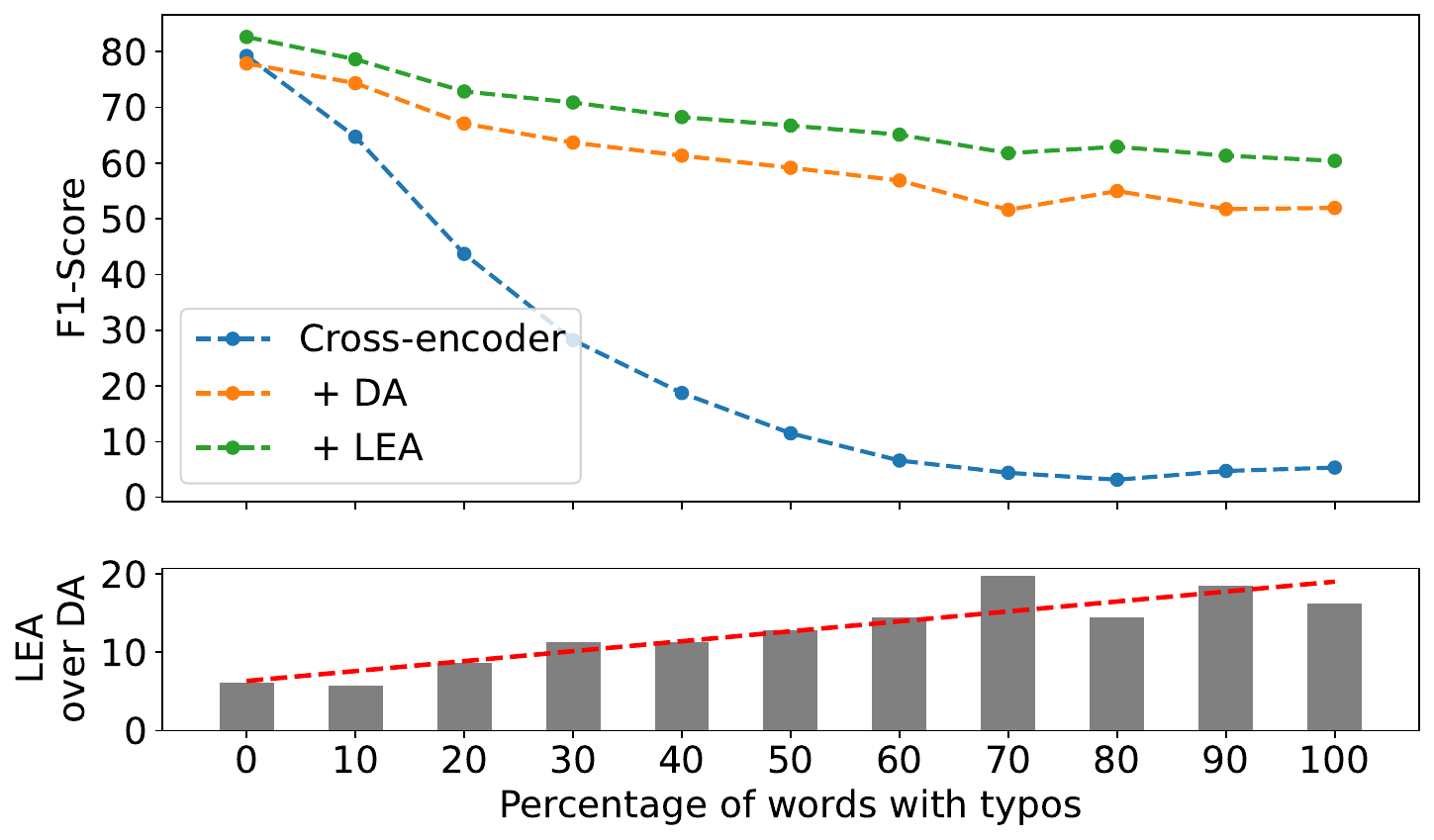}
    \caption{Performance comparison of the cross-encoder approaches under the influence of different noise strengths (top) and the relative improvement of LEA over the cross-encoder trained with data augmentation (bottom). We use BERT-Medium in Abt-Buy for this experiment.}
    \label{fig:noise_gap}
\end{figure}

We analyze in Figure~\ref{fig:noise_gap} (top) the robustness of LEA to different noise strengths at test time.
These results demonstrate a higher robustness to typos than vanilla cross-encoder baselines trained with and without data augmentation. For this experiment, models trained simulating typos use a 20\% probability of introducing them in a word, while at test this probability is modified to change the noise strength.
Intuitively, since the character-level similarities exploited by LEA are not learned during training, they provide the model with 
information that is, to some extent, less dependent on the amount of noise. 
Furthermore, Figure~\ref{fig:noise_gap} (bottom) shows an increasing gap between the performance of LEA with respect to the vanilla cross-encoder trained with data augmentation, suggesting a better generalization of LEA to different noise strengths. 

We, then, can respond $\mathbf{RQ_4}$ based on these results, \textit{LEA maintains a robust performance across noise strengths, being dramatic the performance drop of a vanilla cross-encoder trained without data augmentation}.

\subsection{Additional experiments}

\subsubsection{Comparison with larger models}
\label{sec:appendix_exp_larger_models}

In order to assess the effectiveness of LEA in a larger model, we perform experiments using BERT-Large. Additionally, we adopt auto-regressive architectures (GPT-2 and GPT-Neo) to compare them with the auto-encoder models used across this work.
In Table~\ref{table:appendix_bert_large_typo}, we show that despite following the same training procedure, the gap between the vanilla cross-encoder and LEA using BERT-Large increases to 28 absolute points. In Table~\ref{table:appendix_bert_large_clean} we show the effectiveness of LEA for the clean versions of the datasets.

For the GPT-like models, we followed the same approach as in \cite{2022_arXiv_SGPT} and fine-tuned the backbones as cross-encoders using the last token embedding for the sentence representation pooling (also suggested in~\cite{GPTNeo, GPT2}). 
We used the same hyper-parameters as the ones used in the experiments of our paper, \ie~number of epochs, size of the classification head, etc, and the publicly available pre-trained weights in HuggingFace~\cite{GPTNeo, GPT2}.
As we observe in Table~\ref{table:appendix_gpt_typo} 
, embeddings obtained by fine tuning GPT-like architectures in our downstream tasks still suffer from the textual noise issue, experiencing drops in performance of ~23 and ~7 absolute points on average for GPT2-330M without and with DA, respectively. 
GPTNeo-125M also shows an average drop in performance of ~21 and ~4 absolute points when trained without and with DA, respectively. 
Despite these models being pre-trained in massive data and having more parameters, we show that by just using BERT-Base equipped with LEA we manage to outperform GPT-like architecture in the presence of textual noise. 
Note that we leave the addition of LEA to GPT-like architectures for future work.

{\renewcommand{\arraystretch}{1.3}
\begin{table}[tp]
    \centering
    \begin{tabular}{L{0.06\textwidth} L{0.078\textwidth} L{0.078\textwidth} L{0.08\textwidth} L{0.08\textwidth} N}
    \hline
    & \multirow{2}{*}{\shortstack[l]{Abt-Buy\\(typo)}} & \multirow{2}{*}{\shortstack[l]{Amaz.-\\Goog.\\(typo)}} & \multicolumn{2}{C{0.15\textwidth}}{WDC-Comp. (typo)} & \tabularnewline
    \cline{4-5} &&& Med. & Large & \tabularnewline
    \hline
    \hline
    BERT-L       & 48.10 $\pm$ 2.5 & 19.27 $\pm$ 1.8 & 48.97 $\pm$ 2.3 & 44.29 $\pm$ 0.9 & \tabularnewline
    BERT-L\newline + DA  & 72.09 $\pm$ 3.0 & 48.95 $\pm$ 1.7 & 76.79 $\pm$ 0.3 & 79.98 $\pm$ 0.7 & \tabularnewline
    BERT-L\newline + LEA & \textbf{76.17 $\pm$ 2.5} & \textbf{69.06 $\pm$ 0.7} & \textbf{82.00 $\pm$ 0.9} & \textbf{86.85 $\pm$}\newline \textbf{0.9} & \tabularnewline
    \hline
    \tabularnewline
    \end{tabular}
    \caption{Results for BERT-Large in the test sets with typos.}
    \label{table:appendix_bert_large_typo}
\end{table}
}
{\renewcommand{\arraystretch}{1.3}
\begin{table}[tp]
    \centering
    \begin{tabular}{L{0.06\textwidth} L{0.078\textwidth} L{0.1\textwidth} L{0.06\textwidth} L{0.06\textwidth} N}
    \hline
    & \multirow{2}{*}{\shortstack[l]{Abt-Buy\\(clean)}} & \multirow{2}{*}{\shortstack[l]{Amaz.-Goog.\\(clean)}} & \multicolumn{2}{C{0.12\textwidth}}{WDC-Comp. (clean)} & \tabularnewline
    \cline{4-5} &&& Med. & Large & \tabularnewline
    \hline
    \hline
    BERT-L       & 84.54 & 69.72 & \textbf{88.32} & 90.46 & \tabularnewline
    BERT-L\newline + DA  & 83.22 & 68.41 & 85.20 & 90.31 & \tabularnewline
    BERT-L\newline + LEA & \textbf{85.45} & \textbf{73.66} & 87.82 & \textbf{92.54} & \tabularnewline
    \hline
    \tabularnewline
    \end{tabular}
    \caption{Results for BERT-Large in the clean test sets.}
    \label{table:appendix_bert_large_clean}
\end{table}
}

These results suggest that larger models might reduce the gap to some extent (as depicted in Table~\ref{table:results_product_typo}), but they strongly suffer with textual noise (as shown when comparing results between Table~\ref{table:appendix_bert_large_typo} and Table~\ref{table:appendix_bert_large_clean}). Overall our approach mitigates the impact of noise, while keeping comparable performance using clean text.

{\renewcommand{\arraystretch}{1.3}
\begin{table*}[htp]
    \centering
    \begin{tabular}{L{0.18\textwidth} L{0.11\textwidth} L{0.12\textwidth} L{0.14\textwidth} L{0.14\textwidth} L{0.14\textwidth} N}
    \hline
    & Abt-Buy & Amaz.-Goog. & WDC-Comp Med & WDC-Comp Large & WDC-Comp XL & \tabularnewline
    \hline
    \hline
    \multicolumn{7}{C{0.8\textwidth}}{Typo} \tabularnewline
    \hline
    GPTNeo-125M         & 52.66 $\pm$ 2.2 & 26.21 $\pm$ 2.5 & 61.69 $\pm$ 1.7 & 68.84 $\pm$ 1.4 & 67.00 $\pm$ 2.6 & \tabularnewline
    GPTNeo-125M + DA    & 69.12 $\pm$ 1.2 & 43.64 $\pm$ 2.0 & 76.38 $\pm$ 0.1 & 81.71 $\pm$ 0.5 & 85.10 $\pm$ 0.9 & \tabularnewline
    GPT2-330M           & 64.39 $\pm$ 2.5 & 41.20 $\pm$ 2.1 & 71.12 $\pm$ 0.4 & 69.18 $\pm$ 1.8 & 68.88 $\pm$ 1.4 & \tabularnewline
    GPT2-330M + DA      & 74.76 $\pm$ 3.2 & 57.29 $\pm$ 1.5 & 74.47 $\pm$ 1.2 & 85.51 $\pm$ 0.5 & 87.31 $\pm$ 0.2 & \tabularnewline
    BERT-Base-110M + LEA & \textbf{75.97 $\pm$ 1.4} & \textbf{70.41 $\pm$ 0.9} & \textbf{82.10 $\pm$ 1.7} & \textbf{88.07 $\pm$ 1.0} & \textbf{89.68 $\pm$ 0.1} & \tabularnewline
    \hline
    \multicolumn{7}{C{0.8\textwidth}}{Clean} \tabularnewline
    \hline
    GPTNeo-125M         & 76.26 & 59.00 & 78.87 & 87.66 & 90.37 & \tabularnewline
    GPTNeo-125M + DA    & 77.96 & 55.56 & 79.52 & 86.41 & 89.64 & \tabularnewline
    GPT2-330M           & 85.38 & 67.85 & 84.70 & 90.32 & \textbf{92.89} & \tabularnewline
    GPT2-330M + DA      & \textbf{87.60} & 65.72 & 82.26 & 88.39 & 91.78 & \tabularnewline
    BERT-Base-110M + LEA & 82.66 & \textbf{72.62} & \textbf{86.40} & \textbf{91.04} & 92.60 & \tabularnewline
    \hline
    \tabularnewline
    \end{tabular}
    \caption{Results for GPT-like models in the test sets with and without typos, respectively.}
    \label{table:appendix_gpt_typo}
\end{table*}
}


\subsubsection{Comparison with larger datasets}
\label{sec:appendix_exp_larger_datasets}

We have conducted experiments considering WDC-Computers XLarge (68,461 data points in total for training) and WDC-All (214,661 samples for training) obtaining the results in Table~\ref{table:appendix_wdc_xlarge}.

{\renewcommand{\arraystretch}{1.3}
\begin{table}[ht]
    \centering
    \begin{tabular}{L{0.075\textwidth} L{0.09\textwidth} L{0.062\textwidth} L{0.09\textwidth} L{0.06\textwidth} N}
    \hline
    & \multicolumn{2}{C{0.152\textwidth}}{WDC-Comp. XL} & \multicolumn{2}{C{0.15\textwidth}}{WDC-All XL} & \tabularnewline
    \cline{2-5} & Typo & Clean & Typo & Clean & \tabularnewline
    \hline
    \hline
    Electra-S       & 57.86 $\pm$ 1.3 & \textbf{93.82} & 57.92 $\pm$ 0.3 & \textbf{92.62} & \tabularnewline
    Electra-S\newline + DA  & 84.88 $\pm$ 1.0 & 93.49 & 82.96 $\pm$ 1.0 & 91.12 & \tabularnewline
    Electra-S\newline + LEA & \textbf{87.24 $\pm$ 0.7} & 93.13 & \textbf{85.81 $\pm$ 0.2} & 91.79 & \tabularnewline
    \hline
    BERT-M           & 60.30 $\pm$ 1.0 & 93.40 & 57.31 $\pm$ 0.7 & \textbf{91.81} & \tabularnewline
    BERT-M\newline + DA      & \textbf{88.10 $\pm$ 1.3} & 92.98 & 79.9 $\pm$ 0.6 & 90.75 & \tabularnewline
    BERT-M\newline + LEA     & 87.96 $\pm$ \textbf{0.7} & \textbf{93.45} & \textbf{86.20 $\pm$ 0.6} & 91.73 & \tabularnewline
    \hline
    BERT-B           & 66.57 $\pm$ 1.3 & 93.65 & 65.29 $\pm$ 0.8 & 90.85 & \tabularnewline
    BERT-B\newline + DA      & 87.74 $\pm$ 0.8 & \textbf{92.98} & 82.20 $\pm$ 0.4 & 89.77 & \tabularnewline
    BERT-B\newline + LEA     & \textbf{89.68 $\pm$ 0.1} & 92.6  & \textbf{85.71 $\pm$ 0.8} & \textbf{91.3}  & \tabularnewline
    \hline
    \tabularnewline
    \end{tabular}
    \caption{Results in WDC-Computer XLarge and WDC-All XLarge for three backbones: Electra-Small (Electra-S), BERT-Medium (BERT-M) and BERT-Base (BERT-B).}
    \label{table:appendix_wdc_xlarge}
\end{table}
}

In all the experiments, we show that LEA consistently improves the baselines performance by a significant margin, confirming the effectiveness of our proposal in larger datasets. It is worth mentioning that the average results of ``BERT-M + DA'' for the 3 test splits slightly improves LEA, although with a high standard deviation. Nevertheless, LEA clearly outperforms the baselines in the remaining scenarios.


\section{Conclusions}
\label{sec5:conclusion}
This work proposes LEA, a LExical-aware relative Attention module designed to improve the performance of cross-encoder architectures in sentence similarity tasks. LEA is particularly intended for scenarios with textual noise (\eg~typos) and short texts, where we show that vanilla Transformers drop performance due to tokenization shifts between noisy and clean data. 
In particular, we propose to modify the self-attention module by introducing a lexical bias. This lexical information tackles the tokenization shift by providing a raw character-level similarity that tends to be high for lexically close words, with and without typos. This similarity is independent of the tokenization and does not assume any prior knowledge on the type of noise present in the data.

The results of LEA on five e-commerce datasets using several backbones of varying size demonstrate consistent improvements when dealing with typos over cross-encoder baselines.
We further verify the robustness of LEA against typos in textual entailment and paraphrasing tasks and observe competitive performance despite not being strictly designed for these scenarios.
Moreover, we provide insights to better understand the behaviour of LEA and explore the impact of: 1) different string similarity metrics, 2) introducing the lexical bias at varying subsets of layers and 3) sharing parameters when encoding the lexical similarities. 
Finally, we investigate the generalization to different noise strengths, demonstrating that LEA performs and generalizes better than the vanilla cross-encoder baselines.

\subsection{Limitations and future work}
Despite making no assumption about the type of noise, LEA assumes that lexical similarities between two sentences is a relevant bias for similarity matching. 
It is worth mentioning that in scenarios without typos there is a slight drop in performance (lower than 2 absolute points on average) as reported in Table~\ref{table:results_product_clean} when adding this bias. 
However, in the presence of typos LEA largely outperforms a Vanilla cross-encoder (more than 30 absolute points on average), thus demonstrating that the proposed lexical bias helps in these scenarios. 
LEA is designed for Transformer configurations where two or more sentences are used as part of the input to the model. 
While limited to this specific context, it encompasses a wide-ranging topic within the sentence similarity literature and LEA can be repurposed across different but related domains. 

Future work will focus on improving the use of lexical information on longer texts and better using this bias in clean scenarios. Another interesting research direction is the extension of LEA to bi-encoders with late-interaction techniques.

\bibliographystyle{ACM-Reference-Format}
\balance
\bibliography{LEA_kdd23}

\appendix

\section{Alternatives to the relative attention bias}
\label{sec:appendix_relative_bias}

{\renewcommand{\arraystretch}{1.3}
\begin{table*}[htbp]
    \centering
    \begin{tabular}{L{0.10\textwidth} L{0.1\textwidth} L{0.1\textwidth} L{0.1\textwidth} L{0.1\textwidth} L{0.1\textwidth} L{0.1\textwidth} N}
    \hline
    & Abt-Buy (clean) & Abt-Buy (typos) & Amaz.-Goog. (clean) & Amaz.-Goog. (typos) & WDC-Large (clean) & WDC-Large (typos) & \tabularnewline
    \hline
    \hline
    Learned        & 77.59 & 69.52 $\pm$ 2.0 & 69.76 & 68.09 $\pm$ 0.4 & 91.11 & \textbf{86.76 $\pm$ 0.6} & \tabularnewline
    Fix sin.       & 79.47 & 72.45 $\pm$ 1.6 & 66.23 & 61.07 $\pm$ 0.7 & 90.91 & 86.63 $\pm$ 0.6 & \tabularnewline
    Fix-scale sin. & \textbf{82.31} & \textbf{73.19 $\pm$ 2.0} & \textbf{73.39} & \textbf{69.30 $\pm$ 1.2} & \textbf{91.23} & 86.66 $\pm$ 0.8 & \tabularnewline  
    \hline
    \tabularnewline
    \end{tabular}
    \caption{Performance for BERT-Med. + LEA varying the way of representing lexical similarities for generating the bias.}
    \label{table:appendix_relative_embd}
\end{table*}
}

Inspired by the relative attention bias introduced in~\cite{2017_NeiurIPS_AttentionAllNeed} (``Fix sin.'' as reported in Table~\ref{table:appendix_relative_embd}), LEA introduces a small variation that scales the similarities to be in the range of $[0, 2\pi]$ instead of $[0, 1]$ (``Fix-scale sin.'' in Table~\ref{table:appendix_relative_embd}).
The motivation behind this change is to increase the granularity of the lexical similarities by expanding the domain of the cosine and sine functions. 
This modification yields representations that are better distributed in the space. 
Note that we also considered an additional alternative to the sinusoidal embeddings by using learnable embeddings (``Learned'' in Table~\ref{table:appendix_relative_embd}). 
Apart from showing lower performance, the learnable embeddings add extra parameters and require the discretization of the similarities to map them into embeddings, a step that could potentially lead to information loss.

As presented in Table~\ref{table:appendix_relative_embd}, the use of sinusoidal functions to represent lexical similarities provides improved robustness and flexibility compared to the use of raw lexical similarity values (\eg~jaccard distance) as the bias in the self-attention module. 


\section{Experimental setup details}
\label{sec:appendix_setup_details}

In this section, we show additional details about the hyper-parameters used across all experiments and backbone architectures. Unless otherwise stated we used the following configuration in all experiments and models:

\begin{itemize}
    \item Batch size: 32.
    \item Learning rate scheduler: cosine annealing with warm-up.
    \item Initial learning rate: $5e^{-5}$.
    \item Warm-up: 1.5 epochs.
    \item Number of epochs: 30.
    \item Weight decay: $5e^{-5}$.
    \item Sentence representation: mean pooling of all token embeddings excluding padding tokens.
    \item Classification head: Multi-Layer Perceptron with a Linear layer of size $256$ follow by LayerNorm, Dropout (0.1), GeLU activation and an output Linear layer of size $2$.
\end{itemize}

In Table~\ref{table:appendix_setup}, we show the differences between the parameters used for the experimental setup of the models under comparison. Note that the Vanilla cross-encoder, regardless the backbone architecture, does not use any data augmentation nor lexical attention embedding. For both the cross-encoder with DA and with LEA we use the same data augmentation described in Table~\ref{table:appendix_setup}.

{\renewcommand{\arraystretch}{1.3}
\begin{table}[htbp]
    \centering
    \begin{tabular}{L{0.12\textwidth} L{0.04\textwidth} L{0.25\textwidth} N}
    \hline
    Method & $d_L$ & Data augmentation & \tabularnewline
    \hline
    \hline
    Cross-encoder & - & - & \tabularnewline
    \hline
    ~~+ DA        & - & \multirow{4}{*}{\shortstack[l]{\\Random per-word character\\augmentation, selected from:\\ \qquad- Deletion\\ \qquad- Insertion\\ \qquad- Substitution\\ \qquad- Swapping\\ \qquad- Keyboard\\ Augmentation probability: \\ \qquad-50\% per sentence\\ \qquad-20\% per word}} & \tabularnewline
    \cline{1-2}
    ~~+ LEA       & 128 & & \tabularnewline
    &&& \tabularnewline
    &&& \tabularnewline
    &&& \tabularnewline
    &&& \tabularnewline
    &&& \tabularnewline
    &&& \tabularnewline
    \hline
    \tabularnewline
    \end{tabular}
    \caption{Differences in the configuration of the models used in all experiments regardless of the model architecture. $d_L$ refers to the size of the pairwise lexical attention embedding.}
    \label{table:appendix_setup}
\end{table}
}

\end{document}